\documentclass{article}

\usepackage{multirow} 

\usepackage{arxiv}
\usepackage{color}

\usepackage[utf8]{inputenc} % allow utf-8 input
\usepackage[T1]{fontenc}    % use 8-bit T1 fonts
\usepackage{hyperref}       % hyperlinks
\usepackage{url}            % simple URL typesetting
\usepackage{booktabs}       % professional-quality tables
\usepackage{amsfonts}       % blackboard math symbols
\usepackage{nicefrac}       % compact symbols for 1/2, etc.
\usepackage{microtype}      % microtypography
\usepackage{lipsum}		% Can be removed after putting your text content
\usepackage{graphicx}
\usepackage{natbib}
\usepackage{doi}

\usepackage[skip=2pt,labelfont=bf,font={small,it}]{caption}

\title{Learning Scene Context Without Images}

\author{ \href{}{\includegraphics[scale=0.06]{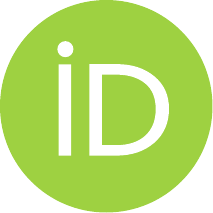}\hspace{1mm}Amirreza Rouhi} \\
        Department of Electrical  \\
        and Computer Engineering,\\        
	Drexel University,\\
	Philadelphia, PA 19104 \\
 \texttt{ar3755@drexel.edu}
	 \\
	\And
	\href{}{\includegraphics[scale=0.06]{orcid.pdf}\hspace{1mm}David Han} \\
	Department of Electrical \\
        and Computer Engineering,\\
	Drexel University,\\
	Philadelphia, PA 19104 \\
	\texttt{dkh42@drexel.edu} \\
}

\date{}

\renewcommand{\headeright}{}
\renewcommand{\undertitle}{ }
\renewcommand{\shorttitle}{LEARNING SCENE CONTEXT WITHOUT IMAGES}

\hypersetup{
pdftitle={A template for the arxiv style},
pdfsubject={q-bio.NC, q-bio.QM},
pdfauthor={David S.~Hippocampus, Elias D.~Striatum},
pdfkeywords={First keyword, Second keyword, More},
}

\begin{document}
\maketitle
\begin{center}
  \captionsetup{type=figure}
 \includegraphics[width=0.45\textwidth]{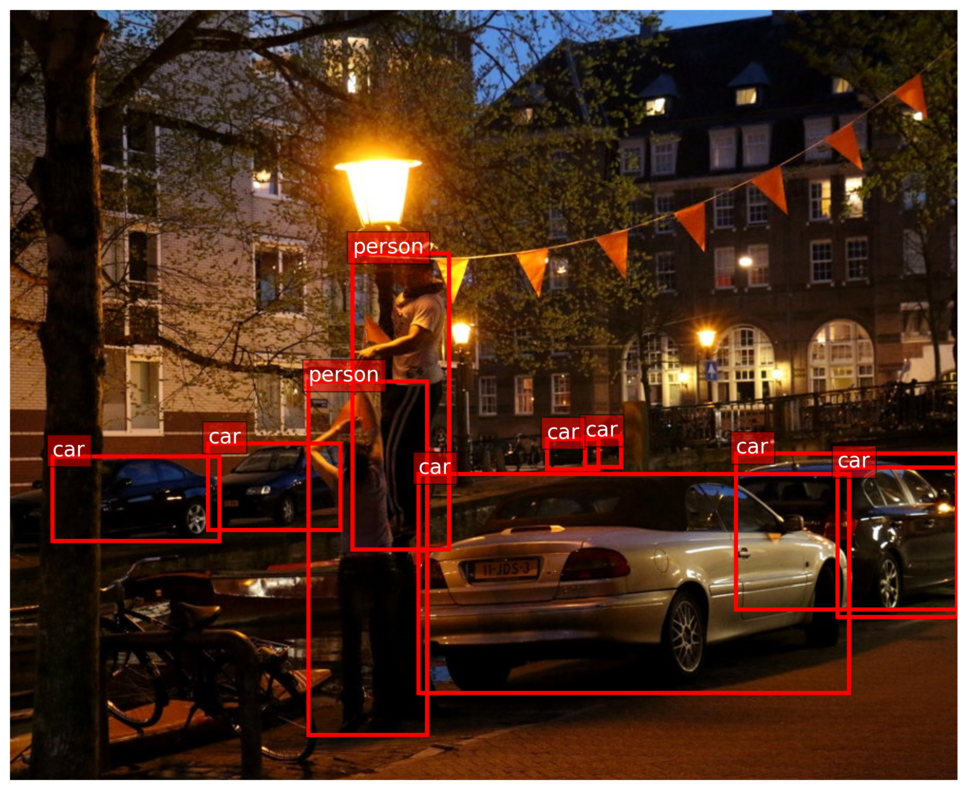}
  \includegraphics[width=0.45\textwidth]{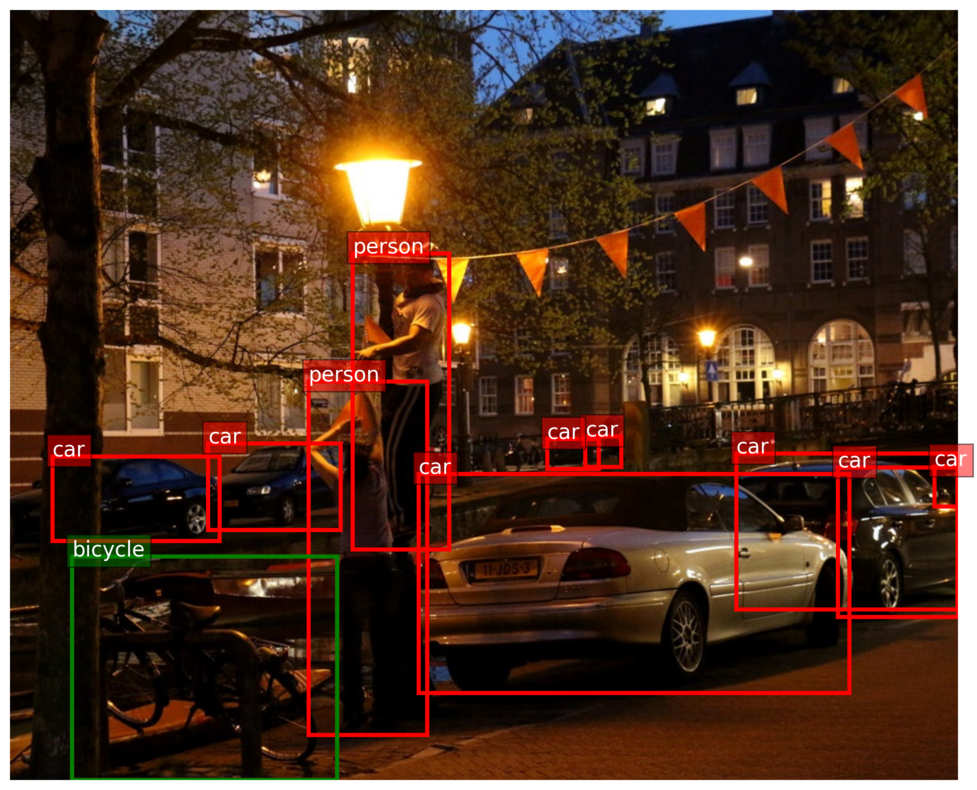}

 %\fbox{\rule[-.5cm]{4cm}{4cm} \rule[-.5cm]{4cm}{0cm}}
 \captionof{figure}{Comparing the Object Detection outcomes using the DETR model with and without our proposed model (LMOD) on a sample from the ExDark dataset \cite{Exdark}. The left image showcases Object Detection without LMOD, while the right image demonstrates Object Detection with LMOD. In the challenging low-light environment, DETR fails to detect the 'bicycle,' whereas the DETR+LMOD integration successfully identifies the 'bicycle' accurately.}

  \label{fig:fig1}
\end{center}
\begin{abstract}
Teaching machines of scene contextual knowledge would enable them to interact more effectively with the environment and to anticipate or predict objects that may not be immediately apparent in their perceptual field. In this paper, we introduce a novel transformer-based approach called $LMOD$ ( Label-based Missing Object Detection) to teach scene contextual knowledge to machines using an attention mechanism. A distinctive aspect of the proposed approach is its reliance solely on labels from image datasets to teach scene context, entirely eliminating the need for the actual image itself. We show how scene-wide relationships among different objects can be learned using a self-attention mechanism. We further show that the contextual knowledge gained from label based learning can enhance performance of other visual based object detection algorithm. 
\end{abstract}

% keywords can be removed

\section{Introduction}

%Consider driving in a poorly lit highway at night.  While what can be discerned on the road is quite limited due to the poor illumination, human drivers, in most cases, detect and identify objects of concern in time and manage to driv safely.  The ability to see and recognize objects in poor visibility is primarily due to our skills to exploit contextual knowledge for anticipating and predicting the type of objects expected to be on the road. In fact, humans apply contextual knowledge to deal with situational uncertainty stemming from incomplete information all the time. Can the same contextual knowledge be taught to machines? As in the case of humans, contextual knowledge would enable machines to predict what can be expected in a scene when full observation is not available to th

%The idea can be further extended for improved machine interaction with objects and entities in a novel environment. While there are many avenues of exploration in teaching machines of contextual knowledge, our interest in this study is primarily focused on visual scenes and the associated context. Our thesis is that by gaining visual situational knowledge, vision algorithms can predict unseen objects as well as improve their recognition performance.
Imagine driving on poorly lit highway at night. Although visibility on the road is severely limited, human drivers, in most cases, adeptly detect and identify objects of concern in a timely manner, ensuring safe navigation. This remarkable ability to perceive and recognize objects in conditions of poor visibility is predominantly due to our ability to leverage contextual knowledge, enabling us to anticipate and predict the types of objects likely to be encountered on the road. Can machines acquire this contextual awareness too? Similar knowledge would empower machines to anticipate scenes even with incomplete data. We believe that visual situational awareness can enhance vision algorithms, enabling them to predict obscured objects and improve recognition.

While deep learning algorithms have made significant progress in computer vision over the past decade, their performance depends on the quality of the image \cite{wang2023context,heo2022occlusion}. 
%recognition capabilities are severely limited when objects are occluded . 

In Figure \ref{fig:fig1}, the left image displays the outcome of object detection using the DEtection TRansformer (DETR) model \cite{carion2020end} on an ExDark dataset \cite{Exdark} known for low-light scenes. DETR fails to detect the bicycle as it only focuses on visible features without considering the  context. We show that coupling a vision algorithm with contextual knowledge enables successful detection, as shown on the right. Remarkably, scene contextual learning is achieved entirely without images. Our contributions are as follows:

- A self-attention model capable of learning contextual scene understanding using only object labels, positions, and sizes.

- A novel integration of our model with a vision algorithm for enhanced object detection in images of poor quality.

\subsection{Related Work}

There has been some efforts to leverage contextual information to enhance object detection. Some methods use Conditional Random Fields or deformable part models to capture object-label relationships, but these approaches struggle in scenarios with missing object regions \cite{rabinovich2007objects, mottaghi2014role}. Others focus on low-level feature statistics but fall short when objects are occluded or small \cite{torralba2003contextual, sun2017seeing}. Transformer-based models like ViT and DETR have made inroads into computer vision, showing promise in contextual scene understanding. However, they predominantly rely on clear visual cues and struggle in complex scenarios with occlusions or small objects \cite{li2022exploring, carion2020end}.

We introduce a novel approach that uses scene context for object prediction, sidestepping the limitations associated with visual features. Our method complements existing vision-based techniques, particularly in challenging scenarios involving low lighting or image blur.

\begin{figure*}
  \centering
\includegraphics[width=0.69\textwidth]{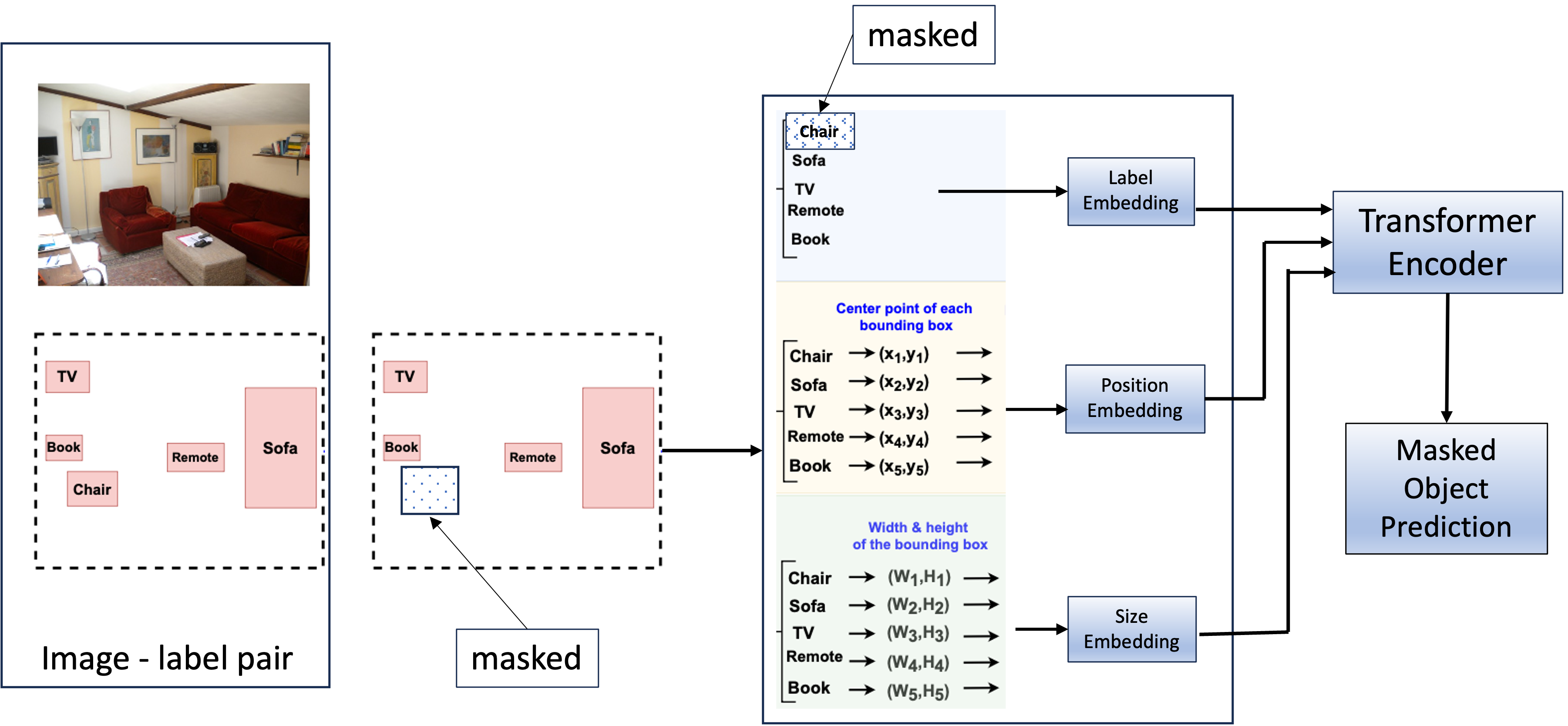}
\caption{LMOD Architecture Overview and Label Mask Training: LMOD takes object class, size, and location information as input. Bounding box details are converted into position and size data and fed into the transformer. Our model utilizes label, size, and position embeddings to predict masked objects, without directly relying on image content.}
 \label{fig:proposed_method_arch}
\end{figure*}
\section{The Proposed Approach}
Our concept revolves around the notion of learning scene context without a reliance on images. We've harnessed the inherent contextual knowledge contained within bounding box annotations in object detection image datasets, such as MS COCO \cite{lin2014microsoft}. Our model, named Label-based Missing Object Detection (LMOD), functions as a context learning network built on a transformer architecture. What sets LMOD apart is its exclusive dependence on bounding box annotations of object labels, bounding box locations, and box sizes, without requiring any pixel-level information, as illustrated in Figure \ref{fig:proposed_method_arch}.  We convert the bounding box information into patch locations and box sizes as depicted in the figure. In a manner similar to training language models, LMOD undergoes self-supervised learning, where portions of the input are masked. However, in contrast to masking words in language models, LMOD masks object category. By employing self-attention mechanisms among the input embeddings of the object categories, their sizes, and their spatial coordinates, the model predicts class category of the masked object. We predict that the training scheme would work equally well when the masking is done on object location or size instead of object label. An investigation into this alternative approach remains as future work. 

\subsection{Input Embeddings}
LMOD model leverages three types of input embeddings: label ($E_L$), position ($E_P$), and size ($E_S$) embeddings. 

\textbf{The category embeddings} capture semantic category through one-hot encoding, WordPiece tokenizer, or Byte-Pair Encoding (BPE) \cite{wu2016google, sennrich2015neural}. These embeddings are organized as an $l \times d$ matrix, where $l$ and $d$ denote the number of unique labels and the dimensionality of the embedding space, respectively.

\textbf{Position embeddings} encode spatial information by segmenting the image into N patches and assigning each object to a patch, $p_i$, based on its bounding box center. We empirically set N=20. 
%The position embedding of each patch, $p_i$, enriches the label embedding of objects located within it.

\textbf{Size embeddings} normalize each object's bounding box area relative to the overall image dimensions. The normalized size is represented as $S^{(object)}=\frac{(x_{br}-x_{tr})*(y_{br}-y_{tl})}{WH}$.

The final object representation, $e^{(i)}$, combines these embeddings: $e^{(i)} = E_L^{(i)} + E_P^{(i)} + E_S^{(i)}$. Each combined embedding is further refined within the Transformer's sub-layers followed by layer normalization.

\subsection{ Masked Label Modeling}
We adapt the Masked Language Modeling (MLM) task from BERT \cite{devlin2018bert} to train our LMOD model using object labels. The MLM loss function is defined as $-\sum_{i=1}^{n}y_{i}\log(\hat{y_{i}})$, where $y_{i}$ and $\hat{y_{i}}$ are the ground truth and predicted labels for the $i$-th object, respectively. Our goal is to minimize this loss by optimizing model parameters. During training, we randomly mask 15\% of the objects to enhance model robustness. We employ three types of embeddings as detailed in Section 2.1, which are concatenated to form the input sequence for the MLM task.

\section{Experiments}

The proposed approach is evaluated in two ways: standalone implementation and integration with image-based object detectors. For the standalone case, the model trained on object embeddings of class, position, and size is tasked to predict missing objects in a scene. In the integrated case, the model is combined with DETR and YOLOv8 object detection models to improve their performance.

\subsection{Label-based Missing Object Detection}
%COCO-2014 has 83K/41K training/validation images across 80 classes.
\textbf{Dataset:} Our experiments leverage two primary datasets, COCO-2014 and VG-500, and their subsets \cite{lin2014microsoft, krishna2017visual}.  We also introduce COCO-80-indoor, a subset with 27,594 training and 13,759 testing images, focused on indoor scenarios. VG-500 is a curated subset of the Visual Genome dataset across the 500 most frequent categories \cite{chen2020knowledge}. Additionally, we employ COCO-1000 \cite{wang2018feedback}, which focuses on the 1,000 most common words extracted from COCO-80.

\textbf{Implementation:} At first, we apply our model for object detection with input consisting only of object labels, and their corresponding positions and relative sizes in the image. We apply the MLM task used in BERT with an object category instead of words. The aim here is similar as in the MLM based training of NLP that the model learns context of the scene by self attention.  We set the learning rate to $5 \times 10^{-5}$ and utilize the AdamW optimizer, which combines the benefits of the Adam optimizer and weight decay regularization.

\textbf{Results:} Given the novelty of our approach, which revolves around predicting masked object categories solely based on detected object annotations, we encountered some challenges in locating comparable models for reference. Although not an exact match, we identified C-TRAN \cite{lanchantin2021general} as a model that shares a related objective of predicting missing labels based exclusively on existing ones. In Table \ref{tab:my-table} we present a comparative analysis of our approach and C-TRAN across three datasets: COCO-80, VG-500, and COCO-1000. It's important to note that this evaluation is carried out using only 50\% of the available labels. While C-TRAN typically relies on image data, the results here reflect its label-only performance in this specific context as reported.

LMOD model outperforms C-TRAN, achieving an average precision (AP) score of 39.8\% compared to C-TRAN's 21.7\% on the COCO-80 dataset when only 50\% of labels are known. Similar trends were observed on VG-500 and COCO-1000 datasets, corroborating the efficacy of our approach.

\textbf{Ablation Study}:  Table \ref{tab:Res-all-emb} presents the results of an ablation study on evaluating different word embedding methods applied to the object categories. Although word embeddings did not yield significant AP gains, byte-level BPE embeddings slightly outperformed others and are thus used in subsequent analyses.
Another ablation study (Table \ref{tab:Ph1-ablation}) examined the roles of position and size embeddings. Position embeddings notably enhanced AP scores by 16.2\% and 15\% on COCO-80-indoor and COCO-80, respectively. In contrast, size embeddings contributed marginal gains. When combined, the AP peaked at 63.3\% and 57.1\% for COCO-80-indoor and COCO-80, validating the utility of these embeddings in object detection.
%\textcolor{red}{Although our model performs commendably on both COCO-80 and COCO-80-indoor datasets, it achieves greater improvement in the indoor scenario, likely owing to reduced complexity in object arrangements.}

% \usepackage{graphicx}
% \usepackage{graphicx}
% \usepackage{graphicx}
\begin{table}[]
    \centering
\caption{The average precision score (\%) of inference with partial labels on three image classification datasets.}
\label{tab:my-table}
\resizebox{0.8\textwidth}{!}{%
\begin{tabular}{|c|c|c|c|}
\hline
Partial Labels Known & \begin{tabular}[c]{@{}c@{}}COCO-80 (50\%)\end{tabular} & \begin{tabular}[c]{@{}c@{}}VG-500 (50\%)\end{tabular} & \begin{tabular}[c]{@{}c@{}}COCO-1000 (50\%)\end{tabular} \\ \hline
C-Tran(no image)     & 21.7                                                   & 24.6                                                  & 27.8                                                      \\ \hline
LMOD                 & \textbf{39.8}                                          & \textbf{32.2}                                         & \textbf{43.5}                                             \\ \hline
\end{tabular}%
}
\end{table}
\begin{table}[]
\centering
\caption{Exploring the Impact of Word Embeddings : A Comparison of Average Precision Scores (\%) }
\label{tab:Res-all-emb}

\resizebox{0.7\textwidth}{!}{%
\begin{tabular}{|c|c|c|c|c|}
\hline
Word Embeddings                                                            & Top-1         & Top-3         & Top-5         & Top-10        \\ \hline
\begin{tabular}[c]{@{}c@{}}LMOD -No Word Emb.\end{tabular}        & 57.0          & 77.6          & 85.1          & 90.5          \\ \hline
\begin{tabular}[c]{@{}c@{}}LMOD -WordPiece Emb.\end{tabular}      & 56.9          & \textbf{77.9} & 85.3          & 90.9          \\ \hline
\begin{tabular}[c]{@{}c@{}}LMOD - BPE Emb.\end{tabular} & \textbf{57.1} & 77.8          & \textbf{85.5} & \textbf{91.0} \\ \hline
\end{tabular}%
}
\end{table}

\begin{table*}[]
    \centering
\caption{Average Precision (\%) of proposed method on COCO-80 and COCO-80-indoor validation sets in different settings}
\label{tab:Ph1-ablation}
\resizebox{\textwidth}{!}{%
\begin{tabular}{|c|cccc|cccc|}
\hline
\multirow{2}{*}{} &
  \multicolumn{4}{c|}{COCO-80-indoor} &
  \multicolumn{4}{c|}{COCO-80 (All Categories)} \\ \cline{2-9} 
 &
  \multicolumn{1}{c|}{Top-1} &
  \multicolumn{1}{c|}{Top-3} &
  \multicolumn{1}{c|}{Top-5} &
  Top-10 &
  \multicolumn{1}{c|}{Top-1} &
  \multicolumn{1}{c|}{Top-3} &
  \multicolumn{1}{c|}{Top-5} &
  Top-10 \\ \hline
LMOD - No Position \& Size Embedding &
  \multicolumn{1}{c|}{42.2} &
  \multicolumn{1}{c|}{68.1} &
  \multicolumn{1}{c|}{72.5} &
  82.4 &
  \multicolumn{1}{c|}{35.3} &
  \multicolumn{1}{c|}{61.1} &
  \multicolumn{1}{c|}{67.5} &
  76.2 \\ \hline
LMOD - Only Size Embedding &
  \multicolumn{1}{c|}{43.1} &
  \multicolumn{1}{c|}{69.4} &
  \multicolumn{1}{c|}{75.1} &
  85.8 &
  \multicolumn{1}{c|}{35.6} &
  \multicolumn{1}{c|}{63.8} &
  \multicolumn{1}{c|}{70.1} &
  79.4 \\ \hline
LMOD - Only Position Embedding &
  \multicolumn{1}{c|}{58.4} &
  \multicolumn{1}{c|}{80.2} &
  \multicolumn{1}{c|}{88.8} &
  92.8 &
  \multicolumn{1}{c|}{50.3} &
  \multicolumn{1}{c|}{74.2} &
  \multicolumn{1}{c|}{83.7} &
  87.5 \\ \hline
LMOD - Position and size Embedding &
  \multicolumn{1}{c|}{\textbf{63.3}} &
  \multicolumn{1}{c|}{\textbf{84.7}} &
  \multicolumn{1}{c|}{\textbf{90.4}} &
  \textbf{96.5} &
  \multicolumn{1}{c|}{\textbf{57.1}} &
  \multicolumn{1}{c|}{\textbf{77.8}} &
  \multicolumn{1}{c|}{\textbf{85.5}} &
  \textbf{91} \\ \hline
\end{tabular}%
}
\end{table*}

\begin{figure*}[hbt!]
  \centering
\includegraphics[width=\textwidth]{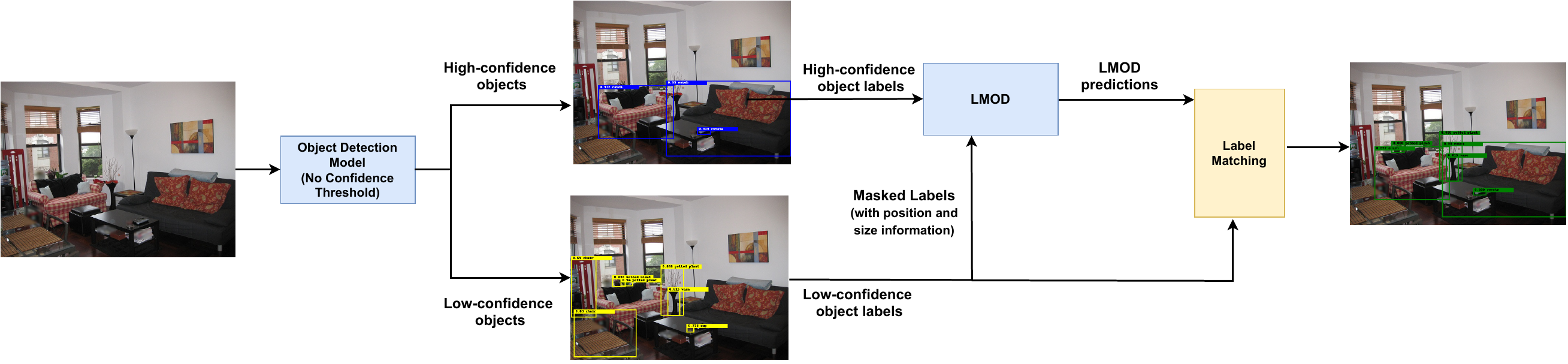}
\caption{Enhancing object detection with LMOD:The figure showcases the steps involved in integrating LMOD for object detection refinement. This includes inputting images into a pre-trained object detection model, dividing predicted labels into "High confidence" and "Low confidence" objects, utilizing LMOD to refine "Low confidence" labels, matching labels based on LMOD suggestions.}
 \label{fig:proposed_method_detr}
\end{figure*}

\begin{table*}[hbt!]
\centering
\caption{Evaluating YOLO8 \& DETR Models with and without LMOD (with $k=5$) Trained on COCO-2017 Dataset and Tested on COCO-2017, COCO-2017-Blurred, and ExDark Datasets: AP (\%) Values by Object Size}
\label{tab:ph2-res}
\resizebox{\textwidth}{!}{%
\begin{tabular}{|c|c|ccc|ccc|ccc|}
\hline
\multirow{2}{*}{} &
  \multirow{2}{*}{Backbone} &
  \multicolumn{3}{c|}{COCO-2017 Validation set} &
  \multicolumn{3}{c|}{COCO-2017-Blured Validation set} &
  \multicolumn{3}{c|}{ExDark} \\ \cline{3-11} 
 &
   &
  \multicolumn{1}{c|}{AP-small} &
  \multicolumn{1}{c|}{AP-medium} &
  AP-large &
  \multicolumn{1}{c|}{AP-small} &
  \multicolumn{1}{c|}{AP-medium} &
  AP-large &
  \multicolumn{1}{c|}{AP-small} &
  \multicolumn{1}{c|}{AP-medium} &
  AP-large \\ \hline
YOLOv8 &
  Darknet-53 &
  \multicolumn{1}{c|}{79.5} &
  \multicolumn{1}{c|}{88.7} &
  89.9 &
  \multicolumn{1}{c|}{70.2} &
  \multicolumn{1}{c|}{82.4} &
  85.2 &
  \multicolumn{1}{c|}{12.0} &
  \multicolumn{1}{c|}{9.7} &
  6.7 \\ \hline
YOLOv8+LMOD &
  Darknet-53 &
  \multicolumn{1}{c|}{\textbf{81.7}} &
  \multicolumn{1}{c|}{\textbf{89.9}} &
  \textbf{90.1} &
  \multicolumn{1}{c|}{\textbf{74.5}} &
  \multicolumn{1}{c|}{\textbf{85.2}} &
  \textbf{86.9} &
  \multicolumn{1}{c|}{\textbf{13.5}} &
  \multicolumn{1}{c|}{\textbf{12.1}} &
  \textbf{7.3} \\ \hline
DETR &
  Resnet101 &
  \multicolumn{1}{c|}{73.9} &
  \multicolumn{1}{c|}{85.4} &
  84.9 &
  \multicolumn{1}{c|}{65.0} &
  \multicolumn{1}{c|}{77.5} &
  80.9 &
  \multicolumn{1}{c|}{5.5} &
  \multicolumn{1}{c|}{3.6} &
  2.4 \\ \hline
DETR + LMOD &
  Resnet101 &
  \multicolumn{1}{c|}{\textbf{75.8}} &
  \multicolumn{1}{c|}{\textbf{86.6}} &
  \textbf{85.2} &
  \multicolumn{1}{c|}{\textbf{68.0}} &
  \multicolumn{1}{c|}{\textbf{78.7}} &
  \textbf{81.1} &
  \multicolumn{1}{c|}{\textbf{6.85}} &
  \multicolumn{1}{c|}{\textbf{4.9}} &
  \textbf{2.8} \\ \hline
\end{tabular}%
}
\end{table*}
\subsection{LMOD Integration in Object Detection}
While LMOD can function as a stand alone model, its utility lies where it is integrated to enhance performance of other models.  One example is its synergy with a pixel-based object detection algorithm like DETR and YOLO. Our demonstration clearly showcased object detection performance enhancement, particularly when dealing with challenging scenarios like low-light conditions, as illustrated in Figure \ref{fig:fig1}. This notable improvement is equally pronounced in cases involving image degradation due to fog or blur. 

The integrated LMOD-DETR model is explained in Figure \ref{fig:proposed_method_detr}. The process unfolds as follows: initially, DETR/YOLO is employed to identify objects within the input image. These detected objects are then divided into two distinct groups. 
% The first group comprises objects with confidence scores surpassing a predetermined over 0.85 for DETR and 0.25 for YOLOv8 which are the default thereshold by that models, while the second group includes objects with confidence scores falling between 0.35 and 0.85 for DETR and 0.05 and 0.25 for YOLOv8. In the typical DETR workflow, objects with high confidence scores are classified as detected, while those in the latter category are disregarded.
The initial category consists of objects with confidence scores exceeding predefined thresholds, specifically 0.85 for DETR and 0.25 for YOLOv8, which are the models' default thresholds. The second category encompasses objects with confidence scores ranging from 0.35 to 0.85 for DETR and from 0.05 to 0.25 for YOLOv8. In the standard workflow of DETR and YOLO, objects with high confidence scores are categorized as detected, while those in the latter group are not considered.
LMOD enters the picture by taking the high-confidence object embeddings as the unmasked input. For the objects within the low-confidence category, their labels are masked, and they are supplied to LMOD alongside the corresponding location and size embeddings, forming another part of the input.

Through the application of self-attention mechanisms among the unmasked and masked object embeddings, LMOD undertakes the task of predicting which objects most likely occupy the locations where object detection model falls below the threshold. When LMOD's predictions align with $top\,k$ DETR/YOLO 's initial predictions of objects with low confidence, the object is classified as detected with the category predicted by LMOD. This approach significantly enhances object detection model's detection capabilities when working with the MS COCO dataset. The advantages of this synergy become particularly evident when dealing with images of poor quality.

%As discussed in the introduction, even the best of visual based object detectors find it challenging to dtect objects that are obscured or blurred. To address the challenge, we propose to integrate LMOD with object detection algorithms to assist in detecting these challenging objects. To demonstrate our concept, we combine LMOD with DETR and Faster RCNN object detection methods

\textbf{Dataset:} We evaluate our model's object detection performance on the COCO-2017 dataset \cite{lin2014microsoft}, with small, medium, and large object categories based on bounding box area. However, one limitation of existing object detection datasets is that most annotated objects are typically visually well-defined and easy to detect. In real-world application of vision algorithm, such guarantees cannot be assured. To add more realism, we introduce COCO-2017-Blurred, a dataset derived from COCO-2017, with randomly blurred one-third of objects using a 21x21 Gaussian kernel. This deliberate introduction of blurred objects simulates the challenging conditions often encountered in practical vision applications, where object visibility is less than ideal. We also used ExDark \cite{Exdark} dataset, comprises 7,363 low-light images spanning 12 categories. 

\textbf{Results:} Table \ref{tab:ph2-res} summarizes the evaluation results for YOLOv8 and DETR trained on the COCO-2017 dataset and tested on COCO-2017, COCO-2017-Blurred, and ExDark. The lower performance on ExDark dataset is expected as the algorithms were never trained on it. The evaluation primarily focuses on precision scores due to the availability of bounding box locations for masked objects in the input.
The results highlight the consistent improvement achieved by integrating LMOD into YOLOv8 and DETR models. There are notable enhancements in AP for small and medium-sized objects, along with improved precision scores for large objects. This underscores the effectiveness of integrating LMOD in enhancing object detection across various object sizes and challenging environments.
%Table \ref{tab:my-table} presents the results of our proposed method on the COCO-2017 dataset and its blurred counterpart. The experiments were conducted using the DETR model with and without the integration of our proposed LMOD approach. Since bounding box locations of masked objects are given in the input, we chose to evaluate the performance based on precision scores instead of mean average precision (mAP). DETR + LMOD achieved higher precision scores compared to the baseline: 75.8\% (small), 86.6\% (medium), and 85.2\% (large). Notably, performance on blurred objects showed a significant improvement, with precision scores of 68.0\%, 78.7\%, and 81.1\%.  \textcolor{red}{In the Exdark dataset, significant improvements are observed across all object sizes, particularly for small and large objects.}
%Faster RCNN + LMOD demonstrated improved precision across all object sizes: 67.6\% (small), 82.1\% (medium), and 81.8\% (large). In the presence of blurring, precision scores further increased to 54.4\%, 72.9\%, and 78.1\%. 
\section{Conclusion}
We introduced LMOD, a novel transformer-based method that learns scene context without a reliance on images. LMOD utilizes size and position embeddings to predict masked objects, relying solely on the categories, sizes, and positions of other objects within the scene. We demonstrated that LMOD can learn scene contextual knowledge by the self-supervised MLM training.  By combining LMOD wother object detection models like DETR or YOLO, we also demonstrated that LMOD can enhance performance of the stand alone DETR or YOLOv8 models in object detection task.  This integrated model showcases a remarkable degree of resilience, even in challenging scenarios like blurred objects and low-light images, attributes that are indispensable for real-world applications.
%In this work, we proposed a novel transformer-based approach to teach the machine context of a scene using attention mechanism and only labels from image datasets. The proposed approach, called LMOD, uses position and size embeddings to forecast masked objects in an image, considering only other objects' labels and their corresponding size and location. We demonstrated that LMOD outperforms existing methods in detecting objects in a wide range of sizes and environments. Our study also included integrating LMOD with image based methods such as DETR or Faster RCNN to enhance detection performance with objects that are obscured or blurred. The results of the integrated approach shows clear performance improvements over those of stand alone DETR/Faster RCNN based detectors. Overall, our novel approach teaches a machine context of scenes using an attention mechanism without any actual image.
%\textcolor{red}{We introduced LMOD, a unique transformer-based method that imparts scene context using attention mechanisms and labels from image datasets. LMOD utilizes size and position embeddings to predict masked objects, relying solely on other objects' labels, sizes, and positions. Our approach outperforms existing methods across diverse sizes and environments, and when integrated with DETR or Faster RCNN, enhances detection, especially for obscured or blurred objects. The integrated approach demonstrates clear performance improvements over standalone detectors. Our innovative method imparts scene context without actual images, advancing object detection capabilities.}

\label{sec:refs}

\bibliographystyle{unsrtnat}
\bibliography{references}  %

\end{document}